\documentclass{llncs}

\usepackage{times}

\usepackage{latexsym} 

\usepackage{graphicx}

\usepackage{xspace}

\title{Allocation in Practice\thanks{The work described here is joint with many colleagues including: Martin Aleksandrov, Haris Aziz, Casey Cahan, Charles Gretton, Phil Kilby, Nick Mattei. NICTA is supported
by the Australian Government through the Department of Communications and the Australian Research Council through the ICT Centre of Excellence Program.
}
 }
\author{Toby Walsh}
\institute{
NICTA and UNSW, Sydney, Australia, email: toby.walsh@nicta.com.au}

\begin{document}

\maketitle

\begin{abstract}
How do we allocate scarce resources? How do we fairly allocate costs?
These are two pressing challenges facing society today. I discuss
two recent projects at NICTA concerning resource and cost allocation.
In the first, we  have been working with
FoodBank Local, a social startup working in collaboration with food bank
charities around the world to optimise the logistics of collecting and
distributing donated food. Before we can distribute this food,
we must decide how to allocate it to different charities and food kitchens.
This gives rise to a fair division problem with several new dimensions,
rarely considered in the literature. In the second, we have been
looking at cost allocation within the distribution network of a large
multinational company. This also has several new dimensions
rarely considered in the literature. 
\end{abstract}

\newcommand{\set}{\mathcal}
\newcommand{\myset}[1]{\ensuremath{\mathcal #1}}

\renewcommand{\theenumii}{\alph{enumii}}
\renewcommand{\theenumiii}{\roman{enumiii}}
\newcommand{\figref}[1]{Figure \ref{#1}}
\newcommand{\tref}[1]{Table \ref{#1}}
\newcommand{\myldots}{\ldots}

\newtheorem{myproblem}{Problem}
\newtheorem{mydefinition}{Definition}
\newtheorem{mytheorem}{Proposition}
\newtheorem{myobservation}{Observation}
\newtheorem{mylemma}{Lemma}
\newtheorem{mytheorem1}{Theorem}
\newcommand{\myproof}{\noindent {\bf Proof:\ \ }}
\newcommand{\myqed}{\mbox{$\Box$}}
\newcommand{\myend}{\mbox{$\clubsuit$}}

\newcommand{\mymod}{\mbox{\rm mod}}
\newcommand{\mymin}{\mbox{\rm min}}
\newcommand{\mymax}{\mbox{\rm max}}
\newcommand{\range}{\mbox{\sc Range}}
\newcommand{\roots}{\mbox{\sc Roots}}
\newcommand{\myiff}{\mbox{\rm iff}}
\newcommand{\alldifferent}{\mbox{\sc AllDifferent}}
\newcommand{\permutation}{\mbox{\sc Permutation}}
\newcommand{\disjoint}{\mbox{\sc Disjoint}}
\newcommand{\cardpath}{\mbox{\sc CardPath}}
\newcommand{\CARDPATH}{\mbox{\sc CardPath}}
\newcommand{\common}{\mbox{\sc Common}}
\newcommand{\uses}{\mbox{\sc Uses}}
\newcommand{\lex}{\mbox{\sc Lex}}
\newcommand{\usedby}{\mbox{\sc UsedBy}}
\newcommand{\nvalue}{\mbox{\sc NValue}}
\newcommand{\slide}{\mbox{\sc CardPath}}
\newcommand{\sliden}{\mbox{\sc AllPath}}
\newcommand{\SLIDE}{\mbox{\sc CardPath}}
\newcommand{\circularslide}{\mbox{\sc CardPath}_{\rm O}}
\newcommand{\among}{\mbox{\sc Among}}
\newcommand{\mysum}{\mbox{\sc MySum}}
\newcommand{\amongseq}{\mbox{\sc AmongSeq}}
\newcommand{\atmost}{\mbox{\sc AtMost}}
\newcommand{\atleast}{\mbox{\sc AtLeast}}
\newcommand{\element}{\mbox{\sc Element}}
\newcommand{\gcc}{\mbox{\sc Gcc}}
\newcommand{\gsc}{\mbox{\sc Gsc}}
\newcommand{\contiguity}{\mbox{\sc Contiguity}}
\newcommand{\PRECEDENCE}{\mbox{\sc Precedence}}
\newcommand{\assignnvalues}{\mbox{\sc Assign\&NValues}}
\newcommand{\linksettobooleans}{\mbox{\sc LinkSet2Booleans}}
\newcommand{\domain}{\mbox{\sc Domain}}
\newcommand{\symalldiff}{\mbox{\sc SymAllDiff}}
\newcommand{\alldiff}{\mbox{\sc AllDiff}}

\newcommand{\slidingsum}{\mbox{\sc SlidingSum}}
\newcommand{\MaxIndex}{\mbox{\sc MaxIndex}}
\newcommand{\REGULAR}{\mbox{\sc Regular}}
\newcommand{\regular}{\mbox{\sc Regular}}
\newcommand{\precedence}{\mbox{\sc Precedence}}
\newcommand{\STRETCH}{\mbox{\sc Stretch}}
\newcommand{\SLIDEOR}{\mbox{\sc SlideOr}}
\newcommand{\NAE}{\mbox{\sc NotAllEqual}}
\newcommand{\mytheta}{\mbox{$\theta_1$}}
\newcommand{\mysigma}{\mbox{$\sigma_2$}}
\newcommand{\mysigmatwo}{\mbox{$\sigma_1$}}

\newcommand{\todo}[1]{{\tt (... #1 ...)}}
\newcommand{\myOmit}[1]{}

\newcommand{\dpsb}{DPSB}
\newcommand{\mathbb}[1]{\mbox{\cal #1}}

\section{Introduction}

The next decade will throw up some fundamental and deep
challenges in resource and cost allocation
that computer science can help solve. 
\begin{description} 
\item[Environmental challenges:]
the world's resources are under increasing
pressure with threats like global warning, 
and with the impact of an increasing population. This 
will require us to find ways to 
allocate scarce resource more efficiently
and more fairly. There will also be increasing
pressure to allocate costs fairly to those
consuming these resources. 
\item[Economic challenges:]
the fall out from the global financial crisis
will continue, with fresh shocks likely
to occur. With growth faltering, both 
government and industry will increasingly
focus on efficiency gains. As wealth concentrates
into the hands of a few, a major and highly topical
concern will be equitability. 
\item[Technological challenges:] new markets enabled by the internet
and mobile devices will emerge.
These markets will require computational mechanisms to
be developed to allocate resources and share costs
fairly and efficiently. 
\end{description}
As an example of one of these new markets, consider users sharing 
some resources in the cloud. How do we design
a mechanism that allocates CPU and memory 
to those users that reflects their different preferences,
and that is fair and efficient? Such a 
mechanism needs to be computational. 
We will want to implement it so that it is
highly responsive and 
runs automatically in the cloud. 
And users will want to implement computational
agents to bid for resources automatically.
As a second example of one of these new markets, 
consider sharing costs
in a smart grid. How do we design
a mechanism that shares costs
amongst users that is fair and 
encourages efficiency? Such a 
mechanism again needs to be computational. 
We will want to implement it so that it is highly responsive
and runs automatically over the smart grid. 
And users will again want to implement computational
agents to monitor and exploit costs in the
smart grid as they change rapidly 
over time.

There are a number of ways in which computation 
can help tackle such resource and cost allocation 
problems.
First, computation can help us set up richer, more realistic
models of resource and cost allocation.
Second, within such computational models, users will be able
to express more complex, realistic, combinatorial preferences
over allocations.
Third, computation can be used to improve efficiency
and equitability, and to explore the trade off between 
the two. And fourth, users will increasingly
farm out decision making to computational agents,
who will need to reason rapidly about how resources
and costs are allocated.

Central to many allocation problems is a trade off 
between equitability and efficiency. 
We can, for example, give each item to the person who
most values it. Whilst this is efficient,
it is unlikely to be equitable. Alternatively, we can
allocate items at random. Whilst this is equitable, 
it is unlikely to maximise
utility. Rather than accept allocations
that are equitable but not efficient, or 
efficient but not equitable, we can now 
use computing power to improve equitability and efficiency. 
Computing 
power can also be used to explore the Pareto frontier between
equitability and efficiency. 

One related area that
has benefitted hugely of late from 
the construction of richer computational models is
auctions. 
Billions of dollars of business have
been facilitated by the development of 
{\em combinatorial} auctions, 
transforming several sectors including
procurement and radio spectrum allocation.
We expect rich new models in resource and cost
allocation will drive similar transformations in
other sectors. It is perhaps not surprising that 
a transformation
has already been seen in auctions where efficiency
(but not equitability) is 
one of the main drivers. By comparison,
many of the examples we give here 
are in the not-for-profit and public
sector where criteria like equitability are
often more important. 
Indeed, equitability is looking increasingly
likely to be a major driver of political and
economic reform over the next few decades. 
In the not-for-profit and public sector, research like 
this can quickly inform both practice and policy
and thereby impact on society in a major way.
Indeed, such richer models are now starting to be 
seen in one area of resource allocation,
namely kidney exchange (e.g. \cite{dpsaamas2012}). 

\section{Allocation in theory}

The theoretical foundations
of resource and cost allocation have been developed
using simple abstract models. For example, one simple model
for resource allocation is ``cake cutting'' in which 
we have a single resource that is infinitely
divisible and agents with 
additive utility functions \cite{cakecut}.
As a second example, a simple model for
cost allocation in cooperative game theory
supposes we can assign a cost
to each subset of agents. 
As we will demonstrate shortly,
abstract models like these ignore the richness and structure of 
many allocation problems met in practice. For example,
allocation problems are often repeated. The problem
we meet today is likely to be similar to the one we will meet 
tomorrow. As a second example, allocation problems are
often online. We must start allocating items before
all the data is available. As a third example,
cost functions are often complex, and dependent on
time and other features of the problem. 
Such real world features offer both a challenge and
an opportunity. For instance, 
by exploiting the repeated
nature of an allocation problem,
we may be able to increase equitability
without decreasing efficiency. On the other hand, 
the online nature of an allocation problem
makes it harder both to be efficient and to be equitable.

Our long term goal then is to develop richer, more realistic 
computational models for resource and
cost allocation, and to design mechanisms 
for such models that can be fielded in practice.
Many of the new applications of such models will be 
in distributed and asynchronous environments 
enabled by the internet and mobile technology.
It is here that computational thinking and computational 
implementation is necessary, and
is set to transform how we fairly and efficiently
allocate costs and resources. Hence, a large
focus of our work is on applying a computational
lens to resource and cost allocation problems. 
To this end, we are concentrating on designing
mechanisms that work well in practice, even in the 
face of fundamental limitations in the worst case.

\section{Allocation in practice}

We have two projects underway in NICTA which
illustrate the richness of allocation problems met in practice.

\subsection{Case study \#1: the food bank problem}


Consider the classical fair division
problem. Fair division problems can be 
categorised along several orthogonal
dimensions: divisible or indivisible 
goods, centralised or decentralised mechanisms,
cardinal or ordinal preferences, 
etc \cite{rasurvey}. Such categories are, however, not
able to capture the richness of a 
practical resource allocation problem
which came to our attention recently. 
\begin{figure}[htb]
\begin{center}
\includegraphics[width=0.6\textwidth]{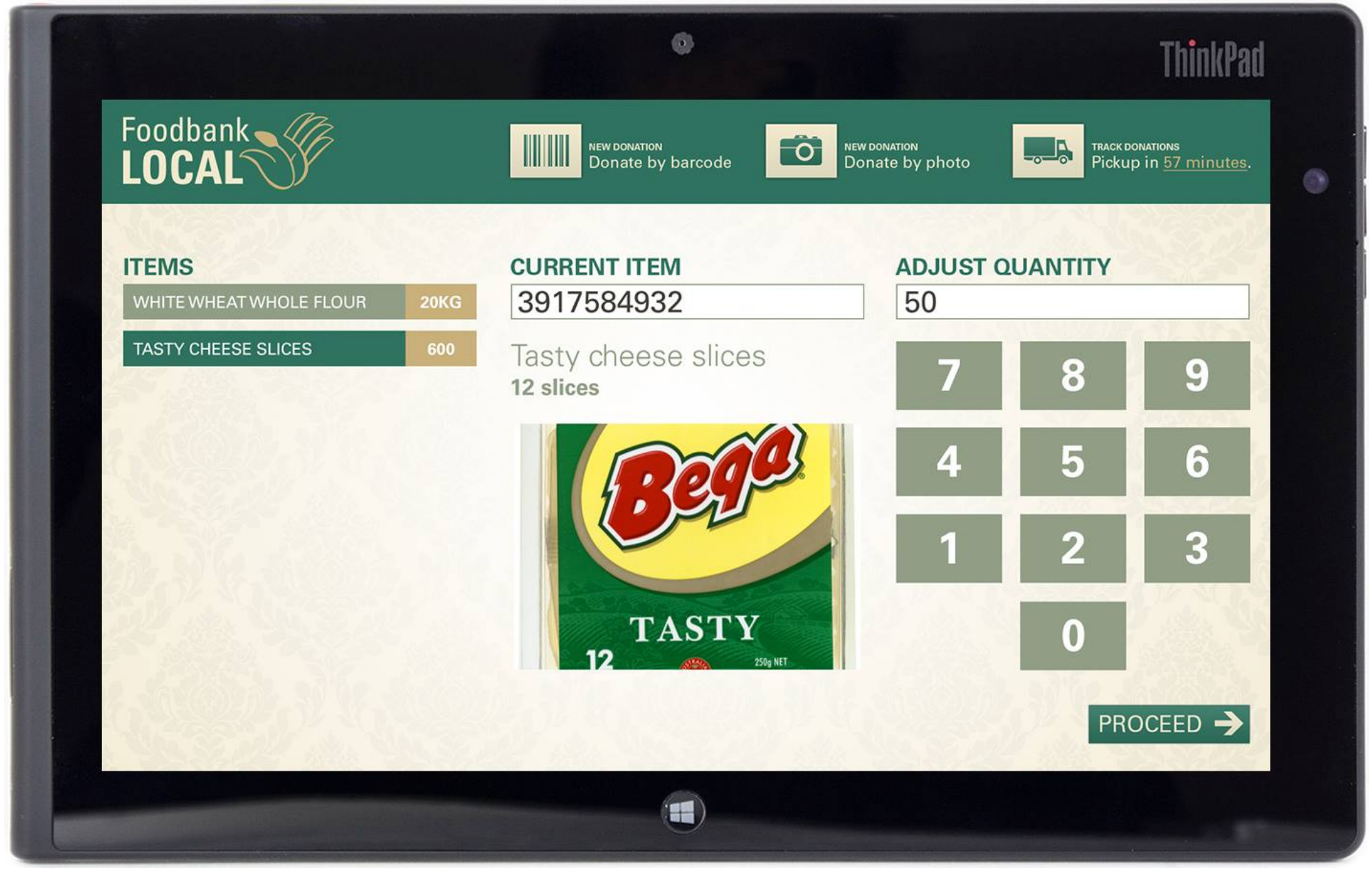}
\end{center}
\caption{FoodBank Local's app for donating, allocating
and distributing food.}
\end{figure}
FoodBank Local is a social startup founded by students from
the University of New South Wales
that is working with food bank charities around
the world, as well as with us,
to improve the efficiency of their operations.
FoodBank Local won the Microsoft Imagine Cup in Australia in 2013,
and were finalists worldwide for their novel
and innovative approach to using technology
for social good. After supermarkets, catering companies
and the public have donated food, the food bank 
must allocate and distribute this food to
charities in the local area. This requires
solving a fair division problem. Fairness
is important as charities often cater to
different sectors of the population (geographical,
social and ethnic), whilst efficiency is
important to maximise the overall good. 
This fair division problem
has several new dimensions, rarely considered
in the literature.
\begin{description} \itemsep=0pt
\item[Online:] Offers of donated food arrive
throughout the day. The food bank cannot wait till
the end of the day before deciding how to
allocate the donated food. This means we have an {\em online} 
problem, where decisions must be made before
all the data is available. 
\item[Repeated:] Each day, a food bank repeats
the task of allocating, collecting and distributing 
food with a similar amount of donated food and set of charities. 
The {\em repeated} nature of the problem provides fresh
opportunities. For example, we can be somewhat
unfair today, in the expectation that we will 
compensate tomorrow. 
\item[Unequal entitlements:] The different charities
working with the food bank have different abilities
at feeding their clients. The allocation of food
needs to reflect this. 
\item[Combinatorial:]
The different charities had complex, combinatorial
preferences over the donated food. 
A charity might want the donated apples or the
bananas but not both. Models based on simple additive
utilities, like those often considered in the literature,
are inadequate to describe their true preferences.
\item[Constrained:]
There are various constraints over the allocations.
For example, we must allocate all the
foods requiring refrigeration to charities
served by the same truck. As a second example,
certain combinations of food cannot be put together
for health and safety or religious reasons. 
\item[Mixed:] 
Each allocation problem induces a new
pickup and delivery problem. This means
that we have a {\em mixed} problem that
combines resource allocation and logistics. 
We need to both ensure a fair division
whilst at the same time optimising distribution. 
\end{description}
To reason about such issues,
we need to develop more complex and realistic
models of resource allocation that borrow
techniques from related areas like constraint
programming \cite{handbookcp}.
Note that some of these features
have individually been considered in the past. 
For example,
we recently initiated the study of online cake cutting
problems \cite{wadt11} in which agents being allocated
the cake arrive over time. In the FoodBank problem, by comparison,
it is the goods that arrive over time. Our model of 
online cake cutting also 
has none of the other features of the FoodBank
problem (e.g. the FoodBank problem is repeated and preferences are
combinatorial, whilst in online cake cutting,
the cake is cut only once and preferences are described
by a simple additive utility function). 
As a second example, Guo, Conitzer and Reeves have looked at 
repeated allocation problems \cite{gcrwine2009} but
their study was limited to just to one indivisible good and
had none of the other features of the FoodBank
problem. 

We now describe our
first step in building a richer model for online fair
division. We stress that this is only the {\em first}
step in putting together a richer model that has
more of the features required by FoodBank Local. 
There are many more features which we need
to add before we have a model that is close
to the actual requirements. 
Our online model supposes items arrive over time and each item
must be allocated as soon as it arrives. For simplicity,
we suppose one new item arrives at each time step. Again, as 
a first step, we also suppose that the utility of
an agent for each item is either 0 or 1. A next
step would be to consider more general utilities.
A simple mechanism for this model is the {\em like}
mechanism. Agents declare whether they like an
item, and we allocate each item uniformly at random
between the agents that like it.

We can study the axiomatic
properties of such an online fair
division mechanism.
For instance, it is strategy proof (the agents
have an incentive to like all items with non-zero
utility and no incentive to like any item with zero
utility) 

\begin{mytheorem}
The like mechanism is strategy proof. 
\end{mytheorem}

The like mechanism is also envy free as an agent 
will not have greater expected utility
for another agent's items. 

\begin{mytheorem}
Supposing agents bid sincerely, the like
mechanism is envy free ex ante. 
\end{mytheorem}

However, ex post,
agents can have considerable envy for 
the actual allocation of items given to another agent. For example, one
agent could get unlucky, lose every random 
draw and end up being allocated no items. 
This agent would then envy any agent allocated
item which they like.
We have therefore been designing more sophisticated
mechanisms which are fairer ex post. A challenge
is to do so without losing a good property
like strategy proofness. 
We also need to explore how
such mechanisms work in practice, and
not just in the worst case. For example,
in real world problems, agents' preferences
are likely to be correlated. A tool
that is likely useful in such studies
is identifying computational phase
transitions (e.g. \cite{isai95,easy-hard,SAT-phase,hardest-SAT,pub702}),
as well as related phenomena
like backbones and backdoors
(e.g. \cite{swijcai2001,kstwaaai2005,kswijcai2005}).

\subsection{Case study \#2: cost allocation in a complex distribution network}

We have come across similar rich features in real world cost 
allocation problems. In particular, we have been working with 
a large multinational company which spends
hundreds of millions of dollars each year
on distributing fast moving consumer goods to 20,000 customers
using a fleet of 600 vehicles. They
face a very challenging problem of allocating costs
between customers\footnote{These are not the actual
costs charged to the customer but the costs used
to decide if a customer is profitable or not. They are
used as the basis for reorganising their business (e.g. 
changing distribution channels, renogiating contracts).}. 
By working with us, they have saved
tens of millions of dollars per annum. 

A standard method to allocate costs is to use co-operative game theory and one of the well defined cost allocation mechanisms like the Shapley value which considers the marginal cost of a customer in every possible subset of customers. In reality, the problem is much richer than imagined in a simple abstract model like the Shapley value which supposes we can simply and easily cost each subset of customers. This richness raises several issues which are rarely considered in the literature. 
Several of these issues are similar to
those encountered in the last example.
For instance, the cost allocation problem is
repeated (every day, we deliver to a similar
set of customers), and constrained (since
we must deliver to all supermarkets of one chain
or none, we must constrain the subsets of customers to
consider only those with all or none of the supermarkets
in the chain). 
However, there are also several new issues to consider: 
\begin{description} 
\item[Computational complexity:] 
we need a cost allocation method that is 
computationally easy to compute. 
As we argue shortly, the Shapley value
is computationally challenging to compute. 
\item[Heterogeneous customers:] 
the Shapley value supposes customers are identical, 
which is not the case in our problem as
different customers order different amounts of
product. In addition, 
the trucks are constrained by volume, weight, and the number
of stops. We therefore need principled methods of
combining the marginal cost of each unit of volume,
of each unit of weight, and of each customer stop. 
\item[Complex cost functions:] a cost allocation
mechanism like the Shapley value supposes
a simple cost function for every subset of customers. 
However, the cost
function in our cost allocation problem 
is much more complex. For example, it is
made up of both fixed and variable
costs. As a second example, the cost function is time
dependent because of traffic whilst 
a mechanism like the Shapley value ignores time. 
As a third example, the cost function 
depends not just on whether we deliver to a
customer or not but on the channel and delivery frequency. 
\item[Strategic behaviour:] 
cooperative game theory supposes the players are
truthful and are not competing with each other. 
In practice, however, we have business alliances
where two or more delivery companies come together to
share truck space to remote areas and must
then share costs. We must consider therefore that the
players may game the system by misrepresenting their 
true goals or costs. 
\item[Sensitivity analysis:] 
in some situations, a small change to the customer 
base has a very large knock-on effect on marginal costs.
We are therefore interested in 
cost allocations that are robust to small changes in the problem. 
\end{description}
As before, some of these features
have been considered individually in the past, but
combinations of these features have
not. For example, \cite{Engevall:2004} 
looked at cost allocation in a vehicle
routing game with a {heterogeneous} fleet of vehicles.
However, this study considered the solution concepts of the core and
the nucleolus rather than the Shapley value. 
The study also ignored other features of the actual real
world problem like
the repeated nature of the delivery problems and
the full complexity of the cost function.

A naive method to allocate costs is simply to use the marginal
cost of each customer. However, such marginal costs 
will tend to under-estimate the actual cost. Consider
two customers in the middle of the outback. Each has a small
marginal cost to visit since we are already visiting the other.
However, their actual cost to the business is half the cost 
of travelling
to the outback. Cost 
allocation mechanism like the Shapley value deal with
such problems. The Shapley value equals the average marginal cost of 
a customer in {\em every} possible subset of customers.
It has nice axiomatic properties like efficiency (it allocates
the whole cost), anonymity (it treats all customers alike)
and monotonicity (when the overall costs go up, no 
individual costs go down). However, we run into several
complications when applying it to the our cost
allocation
problem.

One complication is that the Shapley value 
is computationally challenging to compute in general \cite{CEW11a}.
It involves summing an exponential number of terms (one for
each possible subset of customers), and in our case each term
requires solving to optimality a NP-hard routing
problem. One response to the computational intractability
of computing the Shapley value, is to look to approximate it.
However, we have proved \cite{shapleyvrp} that even finding an
approximation to the Shapley value of a customer
is intractable in general.

\begin{mytheorem}
Unless P=NP, there is no polynomial time $\alpha$-approximation to
the Shapley value of a customer for any constant
factor $\alpha >1$. 
\end{mytheorem}

We have therefore considered heuristic methods based on
Monte Carlo sampling \cite{MaSh60a,approshapley}, 
and on approximating the
cost of the optimal route. 
We have also considered simple proxies to the Shapley
value like the depot distance (that is, allocating 
costs proportional to the distance between customer
and depot). 
Unfortunately we were able to identify pathological
cases where there is no bound on how poorly 
such proxies perform. These pathologies 
illustrate the sort of real world features 
of routing problems like isolated
customers which can cause difficulties. 

\begin{mytheorem}
There exists a $n$ customer problem on which
$\Phi_{depot}/\Phi$ goes to 0 as $n$ goes to $\infty$, and another
on which $\Phi/\Phi_{depot}$ goes to 0
where $\Phi$ is the true
Shapley value for a customer
and $\Phi_{depot}$ is the estimate for it
based on the proxy of distance from depot. 
\end{mytheorem}

Despite such pathological problems, we were able
to show that more sophisticated proxies, especially 
those that ``blend'' together several 
estimates, work well in practice on 
real world data. Figure 1 illustrates the performance
of two different sampling methods. It shows that
in just a few iterations we can converge on
estimates that are within 5-10\%
of the actual Shapley value. 

\begin{figure}[htb]
\begin{center}
\includegraphics[width=0.6\textwidth]{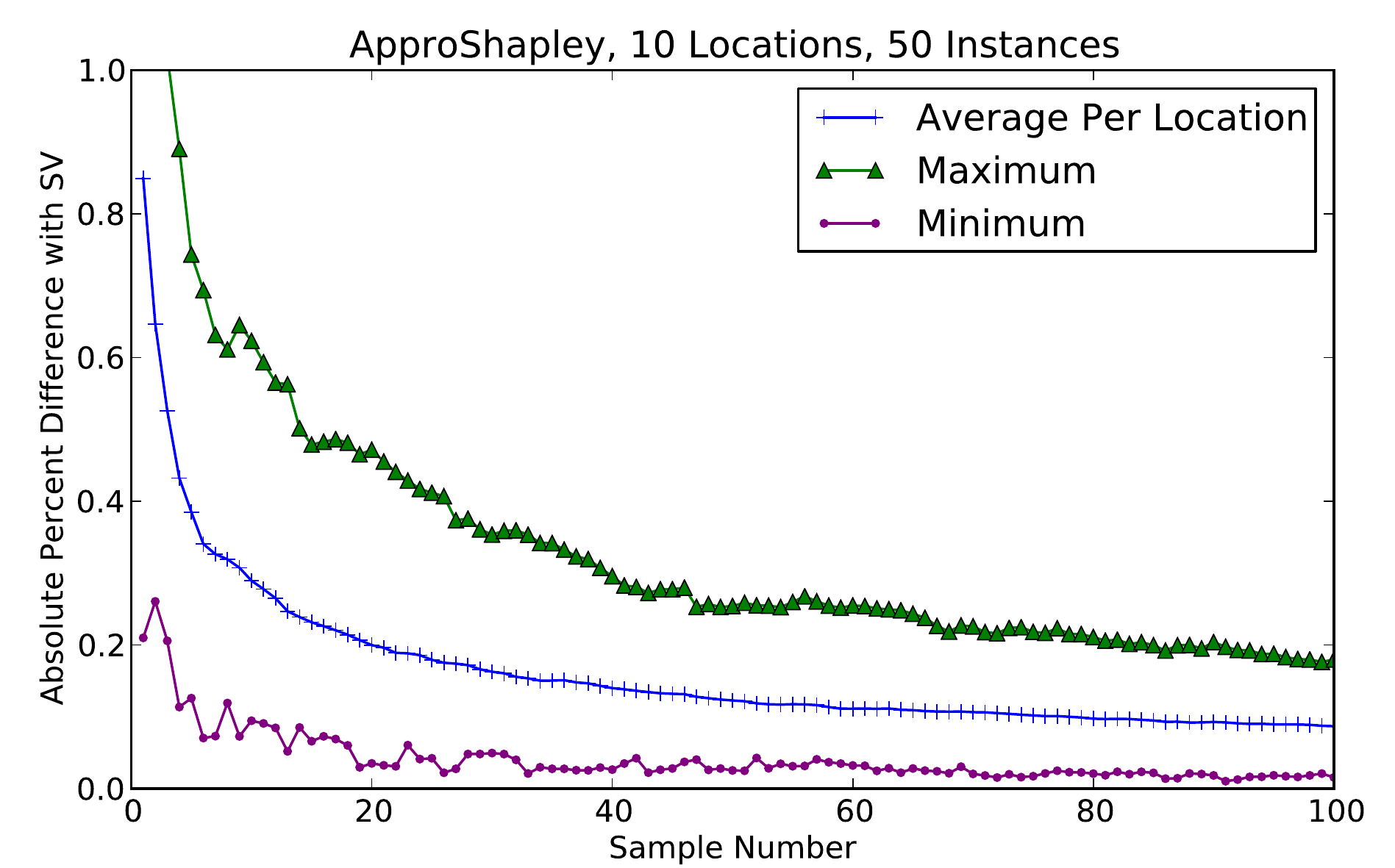}
\end{center}
\caption{Performance of the ApproShapley method 
\cite{approshapley} for Monte Carlo sampling 
to estimate the Shapley value in a distribution costs
game. After just 100 iterations ApproShapley 
achieves an average error of less than 10\% per location 
with a maximum error of less than 20\%.
}
\end{figure}

There are other complications that we
run into when allocating costs in this domain. 
For example, costs are not
simple but made up of fixed and variable
costs which depend on the size of the delivery.
As a second example, costs depend on
the time of day as traffic has a large
impact on the problem.
As a third example, customers should not
be treated equally as they order different
amounts of product, and delivery vehicles are 
limited by both the weight and volume of product that they can
deliver. 
As a fourth example, there are supermarket
chains that require us to deliver to all their
supermarkets or none. We cannot choose some profitable
subset of them. 
All these features need ultimately to be taken 
into account. We are thus only at the beginning
of developing richer models of cost allocation
for this domain. 
Nevertheless, we are encouraged by these preliminary
results which have resulted in significant cost
savings and greatly increased profits for our
business partner.

\section{Related work}

A limited number of richer models for resource and cost
allocation have been considered previously. However, such models
have only considered a small number of the additional
features that we have described here. 

\paragraph{One sided markets:} In an one sided
market, we allocate items to agents based on the
preferences of the agents. The market is one sided
as the items do not have preferences over the
agents. It covers the allocation of goods,
the division of chores, roommates and other related
problems. For example, in the roommates problems,
agents express preferences over each other,
and the goal is to pair these agents up subject
to these preferences. There has been some work already to 
look at more realistic models of resource
allocation in one sided
markets. For example, we have proposed an 
online version of the cake cutting problem
in which agents arrive over time
\cite{wadt11}. As a second example, 
\cite{gcrwine2009} looked at repeated allocation problems 
with a single indivisible good.
As a third example, \cite{bljair2008} studied
the fair division problems of indivisible
goods when agents do not have completely ordered
preferences over the goods, 
but instead have dichotomous and other
succinctly specified types of preferences. 
As a fourth example, 
\cite{imalg2002} considered roommate
problems in which agents can express ties
in their preference lists rather than the basic
assumption of completed ordered preferences.
As a fifth example, \cite{kpsaamas2013}
set up a dynamic version of fair division,
proposed some desirable axiomatic properties
for such dynamic resource allocation, and
designed two mechanisms that satisfy these properties.
However, real world features like those met in these five
examples have usually been
considered in isolation. 
For instance, there has not been proposed a model of 
resource allocation in an one sided market
that is simultaneously online, repeated and involves
preferences that go beyond totally ordered lists. 

\paragraph{Two sided markets:} In a two sided 
market, both sides of the market can have
preferences over each other. For example,
in a stable marriage problem, the women have
preferences over the men they marry and 
vice versa. Another example of a two sided
market is kidney exchange. 
There has been some work already to 
look at more realistic models in two sided
markets. For example, in the hospital residents
problem, a two sided problem in which we allocate
residents to hospitals according to the preferences
of the residents for the hospitals and of the hospitals
for the residents, a number of real
world features have been considered
like couples \cite{mmjco2010},
and ties in preference lists \cite{imsswat2000}.
As a second example, in kidney exchange problems,
a number of real world features have been 
considered like dynamic allocation \cite{dpsaamas2012},
and probabilistic models of clearing (since many
transplants never take place due to unforeseen
logistical and medical problems) \cite{dpsec2013}.
As a third example, the student-project allocation 
problem \cite{aimjda2007} generalises the hospital
residents problem by adding capacity constraints.
However, as with one side markets, such
real world features have usually been
considered in isolation and not in
combination.

\paragraph{Cost allocation:}

Cost allocation has been widely studied in 
game theory and combinatorial optimisation 
\cite{Koster,curiel2008,Young85}.  
Due to its good axiomatic properties, 
the Shapley value has been used
in many domains \cite{Young94}.
A number of special cases have been
identified where it is tractable to
compute (e.g. \cite{MaSh62a}). 
However, due to the computational
challenges we outlined, the Shapley value has rarely been used
in the past in the sort of complex problems like the distribution
problem discussed earlier. An exception is \cite{potters1992}
which introduced the cooperative
travelling salesperson (TSP) game. 
This also introduced the {\em routing game} in which 
the locations visited in a coalition must be traversed
in a (given) fixed order. 
This gives a polynomial time
procedure for computing cost allocations 
\cite{derks:1997}. 
\cite{lundgren1996} developed a column generation  procedure 
to allocate costs for a homogeneous vehicle routing problem. 
\cite{Engevall:2004} extended this to a more practical setting 
of distributing gas using a {\em heterogeneous} fleet of vehicles.
However, this study considered the solution concepts of the core and
the nucleolus rather than the Shapley value. 
It also ignored other real world features like
the repeated nature of such
delivery problems. 
More recently \cite{Ozener:2013} developed  cost allocation
methods for {\em inventory routing} problems
in which customers have a capacity to hold
stock, consume product at a fixed
rate and the goal is to minimise costs
whilst preventing stock-outs. 
Whilst this model has many real world features,
it continues to miss others like inventory and
delivery costs.

\paragraph{Related problems:} 

There have been
a number of complex markets developed to
allocate resources which use money.
For example, in a combinatorial auction,
agents express prices over bundles of items 
\cite{combinatorialauctions}. 
Our two projects, however, only consider allocation
problems where money is not transfered. 
Nevertheless, there are ideas from 
domains like combinatorial auctions which
we may be able to borrow. For example, 
we expect the bidding languages proposed
for combinatorial auction may be useful
for compactly specifying complex, real world
preferences even when money is not being transferred. 
As a second example, as occurs in some course
allocation mechanisms used in practice, we can
give agents virtual ``money'' with which to bid
and thus apply an auction based mechanism
\cite{suier2010,Budish12:Multi}.

Finally,
computational phase transitions have been
observed in a number of related
areas including constraint satisfaction
\cite{gmpwcp95,gmpwaaai96,gmpwaaai97,waaai98,random},
number partitioning
\cite{rnp,gw-ci98},
TSP \cite{tsppt},
social choice
\cite{wijcai09,wecai10,wjair11},
and elsewhere \cite{wijcai99,gwaaai99,ghpwaaai99,wijcai2001,waaai2002}.
We predict that a similar analysis of phase transitions
will provide insight into the precise
relationship between equitability
and efficiency in allocation problems.

\section{Conclusions}

I have discussed 
two recent projects at NICTA involving resource and cost allocation.
Each is an allocation problem with several new dimensions,
rarely considered in the literature. 
For example, our resource allocation problem is online, repeated,
and constrained, whilst our cost allocation problem is
also repeated and constrained, and additionally involves
a complex cost function. 
These projects suggest that models for allocation
problems need to be developed that are richer
and more complex than the abstract models
which have been used in the past to lay the 
theoretical foundations of the field.

\bibliographystyle{splncs}

\end{document}